\documentclass[conference]{IEEEtran}
\IEEEoverridecommandlockouts

\usepackage{soul}
\usepackage{hyperref}
\usepackage[utf8]{inputenc}
\usepackage{graphicx}
\usepackage{amsmath}
\usepackage{booktabs}

\usepackage{microtype}
\usepackage{graphicx}
\usepackage{subfigure}
\usepackage{booktabs} % for professional tables

\usepackage{amsmath,amsfonts,bm}

\usepackage{xcolor}
\usepackage{colortbl}

\def\eqref#1{equation~\ref{#1}}
\def\1{\bm{1}}

\DeclareMathAlphabet{\mathsfit}{\encodingdefault}{\sfdefault}{m}{sl}
\SetMathAlphabet{\mathsfit}{bold}{\encodingdefault}{\sfdefault}{bx}{n}

\newcommand{\R}{\mathbb{R}}

\usepackage{multirow}
\usepackage{paralist}

\usepackage{booktabs} % for professional tables
\usepackage{multicol}
\usepackage{diagbox}

\usepackage[ruled,noend]{algorithm2e}

\SetCommentSty{mycommfont}

\usepackage{here}

\usepackage{amsmath,amssymb,amsfonts,amsbsy,amsfonts,latexsym}
\usepackage{multirow}
\usepackage{makecell}
\usepackage[labelfont=bf,textfont=it,belowskip=0pt,aboveskip=5pt,tableposition=top]{caption}
\usepackage{xcolor}
\usepackage{colortbl}

\definecolor{colorA}{RGB}{189,201,225}
\definecolor{colorB}{RGB}{103,169,207}
\definecolor{colorC}{RGB}{ 28,144,153}
\definecolor{colorD}{RGB}{  1,108, 89}

\newcolumntype{R}{>{\columncolor{gray!40}}r}
\newcolumntype{L}{>{\columncolor{gray!40}}l}
\newcolumntype{C}{>{\columncolor{gray!40}}c}

\usepackage{tabularx,colortbl,xcolor}
\usepackage{multirow}
\usepackage[normalem]{ulem}
\useunder{\uline}{\ul}{}

\usepackage{enumitem}

\usepackage{xparse}% http://ctan.org/pkg/xparse

\DeclareGraphicsExtensions{.pdf,.png}

\SetKwInput{KwInput}{Input}

\usepackage{longtable}
\usepackage{pgfplots}

\NewDocumentCommand{\var}{O{s} m O{}}{%
  \ensuremath{#1_{#2}^{#3}}% add \vphantom{<bizarre sup>}
}
\usepackage{siunitx}

\definecolor{light-gray}{gray}{0.80}

\newcommand\aref{Algorithm \ref}
\newcommand\eref{Eq.~\ref}
\newcommand\fref{Figure~\ref}
\newcommand\tref{Table~\ref}

\newcommand\ha{ \rowcolor{orange!0}}

\newcommand\hc{ \rowcolor{orange!40}}

\newcommand\gc{ \rowcolor{gray!40}}

\def\0{{\bf 0}}

\def\R{{\mathbb R}}

\usepackage{amsthm}

\usepackage{amsthm}

\newtheorem{mytheorem}{Theorem}
\newtheorem{assumption}[mytheorem]{Assumption}
\newtheorem{lemma}[mytheorem]{Lemma}
\newtheorem{remk}[mytheorem]{Remark}

\newcommand{\OURS}{\textsc{HAWQ}\xspace}
\newcommand{\MP}{\xspace\tiny{MP}}

\begin{document}

\title{HAWQ: Hessian AWare Quantization of Neural Networks with Mixed-Precision}
\author{
Zhen Dong$^{*}$\thanks{$^{*}$Equal contribution.}, Zhewei Yao$^{*}$, Amir Gholami$^{*}$, Michael W. Mahoney, Kurt Keutzer\\
University of California, Berkeley\\
{\tt\small \{zhendong, zheweiy, amirgh, mahoneymw, and keutzer\}@berkeley.edu}

}

\maketitle
%\thispagestyle{empty}

%%%%%%%% BODY TEXT

\begin{abstract}
Model size and inference speed/power have become a major challenge
in the deployment of Neural Networks for many applications.
A promising approach to address these problems is quantization.
However, uniformly quantizing a model to ultra low precision leads to significant accuracy degradation. 
A novel solution for this is to use mixed-precision quantization,
as some parts of the network may allow lower precision as compared to other layers.
However, there is no systematic way to determine the precision of different layers.
A brute force approach is not feasible for deep networks, as the search space for mixed-precision is exponential in the number of layers. Another challenge is a similar factorial complexity for determining block-wise fine-tuning order when quantizing the model to a target precision.
Here, we introduce Hessian AWare Quantization (\OURS), a novel second-order quantization method to address these problems.
\OURS allows for the automatic selection of the relative quantization precision of each layer, based on the layer's Hessian spectrum.
Moreover, \OURS provides a deterministic fine-tuning order for quantizing layers, based on second-order information. 
We show the results of our method on Cifar-10 using ResNet20, and on ImageNet using Inception-V3, ResNet50 and SqueezeNext models.
Comparing \OURS with state-of-the-art shows that we can achieve similar/better
accuracy with $8\times$ activation compression ratio on ResNet20, as compared to DNAS~\cite{wu2018mixed}, and up to $1\%$ higher accuracy with up to $14\%$ smaller models on ResNet50 and Inception-V3, compared to recently proposed methods of RVQuant~\cite{park2018value} and HAQ~\cite{wang2018haq}.
Furthermore, we show that we can quantize SqueezeNext to just 1MB model size while achieving above $68\%$ top1 accuracy on ImageNet.
\end{abstract}

\section{Introduction}
\label{sec:intro}

% -----------------------------------------------------------------
\begin{figure*}[!htbp]
\centering

\includegraphics[width=.49\textwidth,trim=0.2in 0in 0.12in 0in, clip]{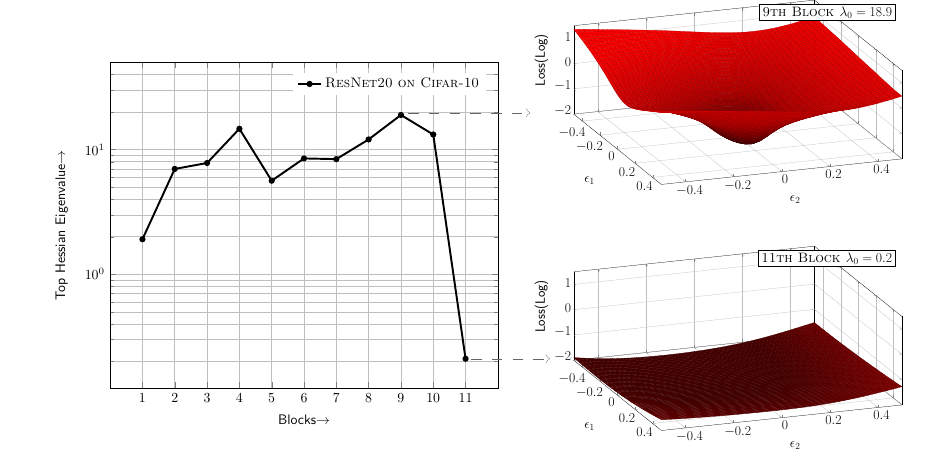}
\includegraphics[width=.49\textwidth,trim=0.2in 0in 0.12in 0in, clip]{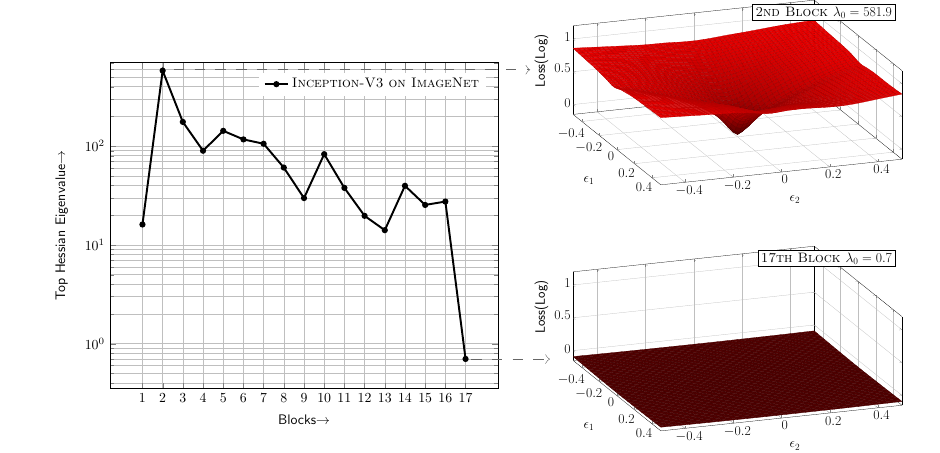}
\caption{
  Top eigenvalue of each individual block of pre-trained ResNet20 on Cifar-10 (Left), and Inception-V3 on ImageNet (Right). Note that the magnitudes of eigenvalues of different blocks varies by orders of magnitude.  See~\fref{fig:resnet20_surface_appendix} and~\ref{fig:inception_surface_appendix} in appendix for the 3D loss landscape of other blocks.
  }
\label{fig:resnet_inception_eigs_surface}
\end{figure*}
% -----------------------------------------------------------------

There has been a significant increase in the computational 
resources required for Neural Network (NN) training and inference.
This is mainly due to larger input sizes (\textit{e.g.}, higher image resolution) as well
as larger NN models requiring more FLOPs and significantly larger memory footprint.
For example, in 1998 the state-of-the-art NN was LeNet-5~\cite{lecun1998gradient} 
applied to MNIST dataset with an input
image size of $1\times28\times28$. Twenty years later, a common benchmark
dataset is ImageNet, with an input resolution that is $200\times$ larger
than MNIST, and with NN models that have orders of magnitude higher memory footprint.

In fact, ImageNet resolution is now considered ``small'' for many applications
such as autonomous driving where input resolutions are significantly larger (more than $40\times$ in certain~cases). 

This combination of larger models and higher resolution
images has created a major challenge in the deployment of NNs
in application environments with computationally constrained resources
such as surveillance systems or ADAS systems in passenger cars. 
This trend is going to accelerate further in the near future.

There has been a significant effort taken by many researchers to address these issues. 
These could be broadly categorized as follows.
(i) Finding  NNs that provide the required accuracy, while remaining compact by design (\textit{i.e.}, with small memory footprint) and which require relatively small FLOPs.
SqueezeNet~\cite{iandola2016squeezenet} was an early effort here, followed by more efficient NNs such as~\cite{sandler2018mobilenetv2, ma2018shufflenet}.
(ii) Co-designing NN architecture and hardware together. 
This can allow significant speed ups and savings in power consumption of the hardware without losing accuracy.
SqueezeNext~\cite{gholami2018squeezenext} is an example work here where the neural network and
associated accelerator are co-designed.
(iii) Pruning redundant filters of NN layers. 
Seminal works here are~\cite{han2015learning, molchanov2016pruning, li2016pruning, mao2017exploring}.
(iv) Applying AutoML for both hardware aware NN design as well as quantization. Notable works
here are DNAS~\cite{wu2018mixed} and HAQ~\cite{wang2018haq}.
(v) Using quantization (reduced precision) instead of float or double precision, which can significantly speed up inference time and reduce power consumption.
This paper exclusively focuses on quantization, but other approaches could be used in conjunction of our method to allow for further possible reduction on the model size.

Quantization needs to be performed for both NN parameters (\textit{i.e.}, weights)
as well as the activations to reduce the total memory footprint of the model
during inference. However, the main challenge here is
that a na\"{\i}ve  quantization can lead to significant loss in accuracy.
In particular, it is not possible to 
reduce the number of bits of all weights/activations of a general convolutional network to ultra low-precision without significant
accuracy loss. 
This is because not all the layers of a convolutional network allow the same quantization
level. 
A possible approach to address this is to
use mixed-precision quantization, where higher precision is used for certain ``sensitive'' 
layers of the network, and lower precision for ``non-sensitive'' layers.
However, the search space for finding the right precision for each layer is
exponential in the number of layers.
Moreover, to avoid accuracy loss we need to perform fine-tuning (\textit{i.e.} re-training) of the model.
As we will discuss below, quantizing the whole model at once and then fine-tuning is not optimal. Instead, we need to perform multi-stage quantization, where
at each stage parts of the network are quantized to low-precision followed by quantization-aware fine-tuning
to recover accuracy. However, the search space to determine which layers to quantize
first is factorial in the number of layers.
In this paper, we propose a Hessian guided approach to address these challenges.
In particular, our contributions are the following.

\begin{enumerate}

    \item The search space for choosing mixed-precision quantization is exponential in the number of layers.
    Thus, we present a novel, deterministic method for determining the relative quantization level of layers based on the Hessian spectrum of each layer.
    \item The search space for quantization-aware fine-tuning of the model is factorial in the number of blocks/layers.
    Thus, we propose
    a Hessian based method to determine fine-tuning order for different NN blocks.
    \item We perform ablation study of \OURS, and we present novel quantization results
    using ResNet20 on Cifar10, as well as Inception-V3/ResNet50/SqueezeNext on ImageNet. Comparison with
    state-of-the-art shows that our method achieves higher precision (up to 1\%), smaller model size (up to $20\%$), and smaller activation size (up to $8\times$).
\end{enumerate}

The paper is organized as follows. 
First, in \S~\ref{sec:related_work}, we will discuss related works on model compression. 
This is followed by describing our method in \S~\ref{sec:methodology}, and our results in \S~\ref{sec:results}.
Finally, we present ablation study in \S~\ref{sec:ablation}, followed by conclusions.

% -----------------------------------------------------------------
\begin{figure*}[!htbp]
\centering
\includegraphics[width=.45\textwidth]{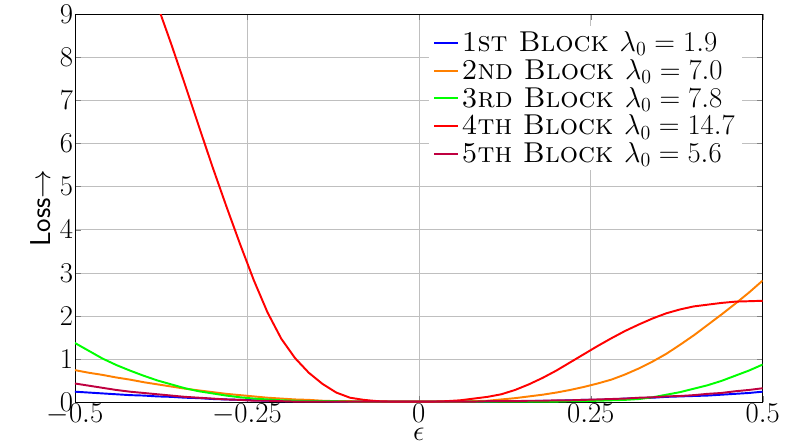}
\includegraphics[width=.45\textwidth]{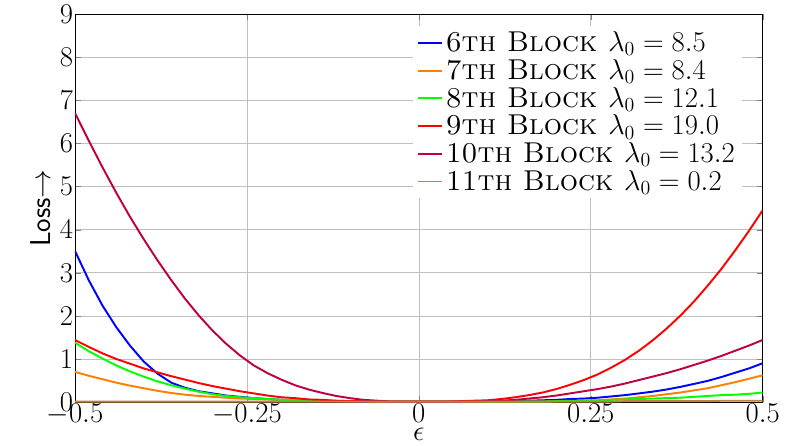}
\caption{
  1-D loss landscape for different blocks of ResNet20 on Cifar-10. The landscape
  is plotted by perturbing model weights along the top Hessian eigenvector of each
  block, with a magnitude of $\epsilon$ (\textit{i.e.}, $\epsilon=0$ corresponds to no perturbation).
  }
\label{fig:resnet_1d_loss_maintext}
\end{figure*}
% -----------------------------------------------------------------

\section{Related work}
\label{sec:related_work}
Recently, significant efforts have been spent on developing new model compression solutions
to reduce the parameter size as well as computational complexity of NNs~\cite{courbariaux2015binaryconnect, han2015deep, hinton2015distilling, rastegari2016xnor, denton2014exploiting, zhou2017incremental, krishnamoorthi2018quantizing, howard2017mobilenets, chollet2017xception, zhang2018shufflenet}. In~\cite{han2015learning, mao2017exploring, li2016pruning}, pruning is used to reduce the number of non-zero weights in NN models. 
This approach is very useful for models that have very large fully connected layers (such as AlexNet~\cite{krizhevsky2012imagenet} or VGG~\cite{simonyan2014very}). 

For instance, the first fully-connected layer in VGG-16 occupies 408MB alone, which is 77.3\% of total
model size. Large fully-connected layers have been removed in other fully convolutional networks such as ResNet~\cite{he2016deep}, or Inception family~\cite{szegedy2016rethinking}.

Knowledge distillation introduced in~\cite{hinton2015distilling} is another direction for compressing NNs.
The main idea is to distill information from a pre-trained, large model into a smaller model.
For instance, it was shown that with knowledge distillation it is possible
to reduce model size by a factor of $3.6$ with an accuracy of $91.61\%$ on Cifar-10~\cite{romero2014fitnets}.

Another fundamental approach has been to architect models which are, by design, both small and hardware-efficient. An initial effort here was 
SqueezeNet~\cite{iandola2016squeezenet} which could achieve
AlexNet level accuracy with $50\times$ smaller footprint through network design, and additional $10\times$ reduction
through quantization~\cite{han2015deep}, 
resulting in a NN with $500\times$ smaller memory footprint. 
Other notable works here are~\cite{howard2017mobilenets, sandler2018mobilenetv2, zhang2018shufflenet, ma2018shufflenet, chollet2017xception}, where more accurate networks are presented. 
Another work here is SqueezeNext~\cite{gholami2018squeezenext}, where a similar
approach is taken, but with co-design of both hardware architecture along with a compact NN model.

Quantization~\cite{asanovic1991experimental,courbariaux2015binaryconnect, rastegari2016xnor, zhou2017incremental, zhou2016dorefa, choi2018pact, Zhang_2018_ECCV} is another orthogonal approach for model compression, where lower bit representation are used instead of redesigning the NN. 
One of the major benefits of quantization is that it increases a NN's arithmetic intensity (which
is the ratio of FLOPs to memory accesses). This is particularly helpful for layers that are memory bound and have
low arithmetic intensity. After quantization, the volume of memory accesses reduces, which can alleviate/remove the
memory bottleneck.

However, directly quantizing NNs to ultra low precision may cause significant accuracy degradation. 

One possibility to address this is to use Mixed-Precision quantization (MP)~\cite{wu2018mixed, zhou2018adaptive}. A second possibility, Multi-Stage Quantization (MSQ), is proposed by~\cite{zhou2017incremental, dong2017learning}. MP and MSQ can improve the accuracy of quantized NNs, but face an exponentially large search space.
This is a major problem that has not been addressed in existing literature for quantization.
Applying existing methods require often ad-hoc rules to choose precision of different layers which
are problem/model specific and do not generalize. The goal of our work here is to address this challenge using second-order
information.

\section{Methodology}
\label{sec:methodology}

Assume that the NN is partitioned into $b$ blocks denoted by $\{B_1,~B_2~\ldots,~B_{b}\}$, with  
learnable parameters $\{W_1,~W_2,~\ldots,~W_{b}\}$.
A block can be a single/multiple layer(s) (or a single/multiple residual block(s) for the case of residual networks). 
For a supervised learning framework, the loss function $L(\theta)$ is:
% \small
\begin{equation}
\label{eqn:basic_problem}
L(\theta) = \frac{1}{N} \sum_{i=1}^{N} l(x_i, y_i, \theta),
\end{equation}
% \normalsize
where $\theta\in\R^d$ is the combination of $\{W_1,~W_2,~\ldots,~W_{b}\}$,
and $l(x, y, \theta)$ is the loss for a datum $(x, y) \in (X, Y)$. 
Here, $X$ is the input set, $Y$ is the corresponding label set, and $N=|X|$ is the size of the training set.

The training is performed by solving an Empirical Risk Minimization problem, to find the optimal model parameters.
This process is typically performed in single precision, where
both the weights and activations are stored with 32-bit precision.

After the training is finished, each of these blocks will have a specific
distribution of floating point numbers for both the parameters, $\theta$, as well as input/output activations.
For quantization, we need to restrict
these floating numbers to a finite set of values, defined by the following function:
\begin{equation}
\label{eqn:quantization_function}
    Q(z) = q_j, \quad \mbox{for}~z \in (t_j, t_{j+1}],
\end{equation}
where $(t_j, t_{j+1}]$ denotes an interval in the real numbers ($j=0,~\ldots~,2^k-1$),
$k$ is the quantization bits,
and $z$ is either an activation or the weights.
This means that all the values in the range of $(t_j, t_{j+1}]$ 
are mapped to $q_j$.
In the extreme case
of binary quantization ($k=1$), $Q(z)$ is basically the sign function.
For cases other than binary quantization, the choice of 
these intervals can be important. One popular option is to
use a uniform quantization function, where the above range
is equally split~\cite{zhou2016dorefa,hubara2017quantized}.
However, it has been argued that
(i) not all layers have the same distribution of floating point values, and
(ii) the network can have significantly different \emph{sensitivity} to quantization of each layer.
To address the first issue, different quantization schemes such as uniformly discretizing logarithmic-domain
have been proposed~\cite{miyashita2016convolutional}.
However, this does not completely address the sensitivity problem.
A sensitive layer cannot be quantized to
the same level as a non-sensitive~layer.

One possible approach that can be used to measure quantization sensitivity is to use first-order information, based on the gradient vector. 
However, the gradient can be very misleading. 
This can be easily illustrated by considering a simple 1-d parabolic function of the form $y=\frac{1}{2}ax^2$ at origin (\textit{i.e.}, $x=0$).
The gradient signal at the origin is zero, irrespective of the value of $a$. 
However, this does not mean that the function is not sensitive to perturbation in $x$.  
We can get a better metrics for sensitivity by using second-order information, based on the Hessian matrix.
This clearly shows that higher values of $a$ result in more sensitivity to input perturbations. 

For the case of high dimensions, the second order information is stored in the Hessian matrix, of size $n_i\times n_i$ for each block. 
For this case, we can compute the eigenvalues of the Hessian to measure sensitivity, as described next.

\subsection{Second-Order Information}
We compute the eigenvalues of the Hessian (\textit{i.e.}, the second-order operator)
of each block in the network. Note that it is not possible to explicitly
form the Hessian since the size of a block (denoted by $n_i$ for i$^{th}$ block) can be quite large.
However, it is possible to compute the Hessian eigenvalues without explicitly forming
it, using a matrix-free power iteration algorithm~\cite{yao2018hessian,martens2010deep,yao2018large}.
This method requires computation of the so-called Hessian \emph{matvec}, which
is the result of multiplication of the Hessian matrix with a given (possibly random)
vector $v$. 
To illustrate how this can be done for a deep network, let us first denote $g_i$ as the gradient of loss $L$ with respect to the $i^{th}$ block parameters, 

\begin{equation}
g_i = \frac{\partial L}{\partial W_i}.
\end{equation}

For a random vector $v$ (which has the same dimension as $g_i$), we have:
\begin{equation}\label{eqn:hessian_free}
\frac{\partial(g_i^Tv)}{\partial W_i} = \frac{\partial g_i^T}{\partial W_i}v +  g_i^T\frac{\partial v}{\partial W_i} = \frac{\partial g_i^T}{\partial W_i}v = H_iv,
\end{equation}
where $H_i$ is the Hessian matrix of $L$ with respect to $W_i$. 
Note that the second equality above, comes from the fact that
$v$ is independent of $W_i$.
We can then use power-iteration method to compute the top eigenvalue of $H_i$,
as shown in~\aref{alg:power_iteration}. 
Intuitively the algorithm requires multiple
evaluations of the Hessian matvec, which can be computed using~\eref{eqn:hessian_free}.

\begin{algorithm}[t]
\DontPrintSemicolon
\caption{Power Iteration for Eigenvalue Computation}
\label{alg:power_iteration}
    \SetAlgoLined
    \KwInput{Block Parameter: $W_i$.
    }
    
    Compute the gradient of $W_i$ by backpropagation, \emph{i.e.}, $g_i=\frac{d L}{d W_i}$.
    
    Draw a random vector $v$  (same dimension as $W_i$).
    
    Normalize $v$, $v=\frac{v}{\|v\|_2}$
    
    \For(\ \ \quad \quad\quad\quad\quad\tcp*[h]{Power Iteration}){i $=1,2,\ldots, n$}{
        Compute $gv = g_i^Tv$ \tcp*{Inner product}
        
        Compute $Hv$ by backpropagation, $Hv = \frac{d(gv)}{dW_i}$ \tcp*{Get Hessian vector product}
        
        Normalize and reset $v$, $v = \frac{Hv}{\|Hv\|_2}$
    }
\end{algorithm}

It is well known, based on the theory of Minimum Description Length (MDL), that fewer
bits are required to specify a flat region up to a given threshold,
and vice versa for a region with sharp curvature~\cite{rissanen1978modeling,hochreiter1997flat}.
The intuition for this is that the noise created by imprecise location of
a flat region is not magnified for a flat region, making it more amenable to aggressive quantization.
The opposite is true for sharp regions, in that even small round off errors
may be amplified. Therefore, it is expected that layers with higher Hessian spectrum (\emph{i.e.}, larger eigenvalues) are more sensitive to quantization.
The distribution of these eigenvalues for different blocks are shown in~\fref{fig:resnet_inception_eigs_surface} for ResNet20 on CIFAR-10 and
Inception-V3 on ImageNet. As one can see, different blocks
exhibit orders of magnitude difference in the Hessian spectrum. For instance,
ResNet20 is an order of magnitude more sensitive to perturbations to its $9^{th}$ block, than its last block.

\begin{figure*}[!htbp]
\centering
\includegraphics[width=.45\textwidth]{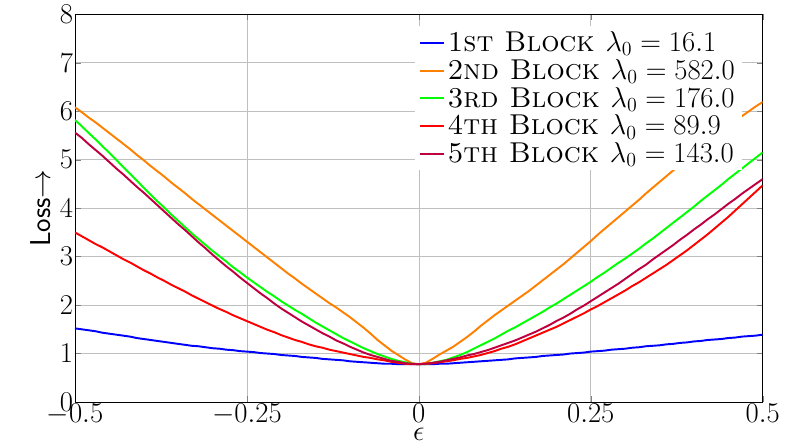}
\includegraphics[width=.45\textwidth]{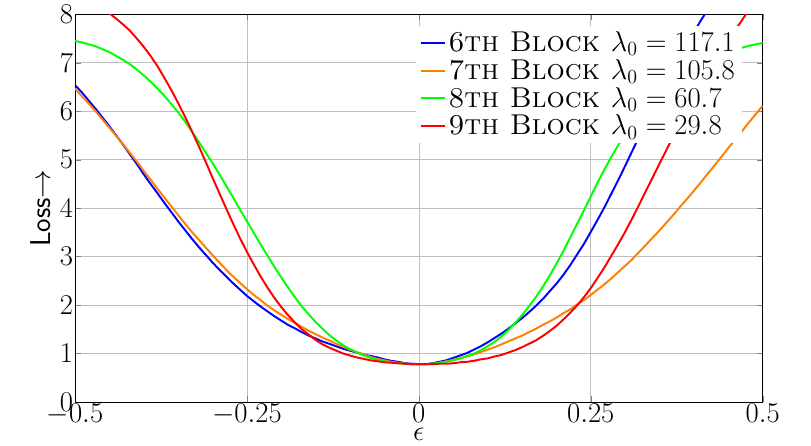}
\includegraphics[width=.45\textwidth]{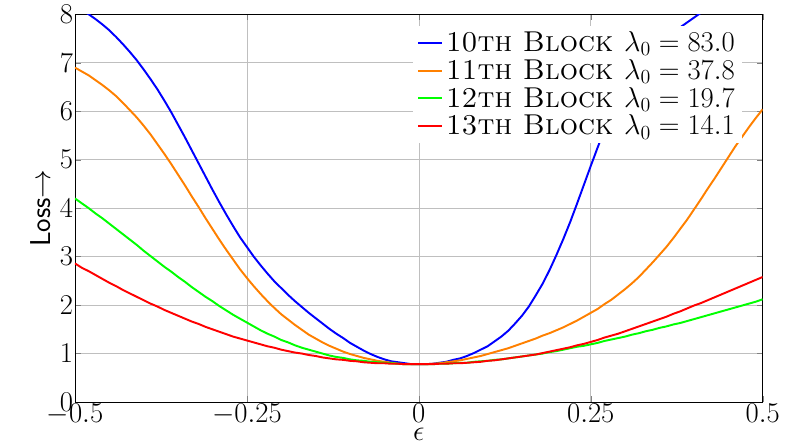}
\includegraphics[width=.45\textwidth]{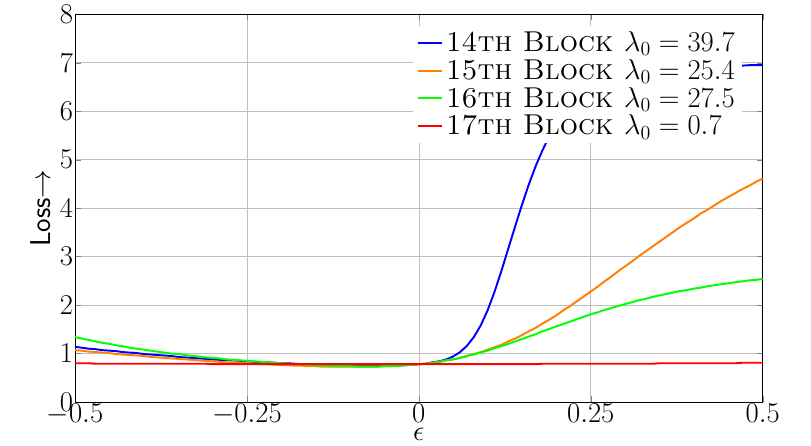}
\caption{
  1-D loss landscape of all blocks of Inception-V3 on ImageNet along the first dominant eigenvector of the Hessian. Here $\epsilon$ is the scalar that perturbs the parameters of the corresponding block along the first dominant eigenvectors. 
  }
\label{fig:inception_1d_loss_appendix}
\end{figure*}

To further illustrate this, we provide 1D visualizations of the loss landscape as well. 
To this end, we first compute the Hessian eigenvector of each block, and
we perturb each block individually along the eigenvector and compute how the loss changes.
This is illustrated in~\fref{fig:resnet_1d_loss_maintext} and~\ref{fig:inception_1d_loss_appendix} for ResNet20 (on Cifar-10) and Inception-V3 (on ImageNet), respectively. 
It can be clearly seen that blocks with larger Hessian eigenvalue (\textit{i.e.}, sharper curvature) exhibit 
larger fluctuations in the loss, as compared to those with smaller Hessian eigenvalue (\textit{i.e.}, flatter curvature). A corresponding 3D plot is also shown in~\fref{fig:resnet_inception_eigs_surface}, where instead of just considering the
top eigenvector, we also compute the second top eigenvector and visualize
the loss by perturbing the weights along these two directions. These surface plots are computed for the
$9^{th}$ and last blocks of ResNet20, as well as 2nd and last blocks of Inception-V3 (the loss landscape for other blocks is shown in~\fref{fig:resnet20_surface_appendix} and~\fref{fig:inception_surface_appendix}).

% --------------------------------------------------------------------
% --------------------------------------------------------------------

% \subsection{Preliminary}

\begin{algorithm}[t]
\DontPrintSemicolon
\caption{Hessian AWare Quantization}
\label{alg:hq}
    \SetAlgoLined
    \KwInput{Block-wise Hessian eigenvalues $\lambda_i$ (computed from~\aref{alg:power_iteration}), and block size $n_i$ 
    for $i=1,\cdots,b$.

    }

    \For(\tcp*[h]{Compute Quantization Precision}){i $=1,2,\ldots, b$}{

        $S_i = \lambda_i/n_i$ \tcp*{See~\eref{eqn:define_S}}
    }

    Order $S_i$ in descending order and to determine relative quantization precision for each block.
    
    Compute $\Delta W_i$ based on~\eref{eqn:quantization_function}.

    \For(\quad \quad \quad\quad\quad \tcp*[h]{Fine-Tuning Order}){i $=1,2,\ldots, b$}{
       $\Omega_i = \lambda_i\|\Delta W_i\|^2$ \tcp*{See~\eref{eqn:define_O}}
    }
    
    Order $\Omega_i$ in descending order and perform block-wise fine-tuning
\end{algorithm}

\subsection{Algorithm}
We approximate the Hessian as a block diagonal matrix, scaled by
its top eigenvalue, $\lambda$, as $\{H_i\approx \lambda_i I\}_{i=1}^{b}$.
Based on the MDL theory, layers with large $\lambda$
cannot be quantized to ultra low precision without significant perturbation to the model.
Thus we can use the Hessian spectrum of each block to sort the different blocks and perform
less aggressive quantization to layers with large spectrum. However, some
of these blocks may contain very large number of parameters, and using higher bits here
would lead to large memory footprint of the quantized network.
Therefore, as a compromise, we weight
the spectrum with block's memory footprint and use the following metric for sorting the blocks:
\begin{equation}\label{eqn:define_S}
S_i = \lambda_i/n_i,
\end{equation}
where $\lambda_i$ is the top eigenvalue of $H_i$.
Based on this sorting, layers that have large number of parameters and have small eigenvalue would
be quantized to lower bits, and vice versa.
That is, after $S_i$ is computed, we sort $S_i$ in descending order and use it as a metric
to determine the quantization precision.%
\footnote{Note that, as mentioned in the limitations section, $S_i$ does not give us the exact bit precision
but a relative ordering for the bits of different blocks.}

Quantization-aware re-training of the neural network is necessary to recover performance which can sharply drop due to ultra-low precision quantization. A straightforward way to do this is to re-train (hereafter referred to as fine-tune) the whole quantized network at once. However, as we will discuss
in \S\ref{sec:results}, this can lead to sub-optimal results. A better strategy
is to perform multi-stage fine-tuning. However, the order in multi-stage tuning is important
and different ordering could lead to very different accuracies.

We sort different blocks for fine-tuning based on the following metric:
\begin{equation}\label{eqn:define_O}
    \Omega_i = \lambda_i \|Q(W_i)-W_i\|_2^2,
\end{equation}
where $i$ refers to $i^{th}$ block, $\lambda_i$ is the Hessian eigenvalue, and $\|Q(W_i)-W_i\|_2$ is the
$L_2$ norm of quantization perturbation.
The intuition here is to first fine-tune layers that have high curvature as
well as large number of parameters which cause more perturbations after quantization.
Note that the latter metric depends on the bits used for quantization and is
not a fixed metric. 
(See~\tref{tab:shift_table} in appendix, where we show how this metric changes for different quantization precision.)
The motivation for choosing this order is that fine-tuning blocks
with large $\Omega_i$ can significantly affect other blocks,
thus making prior fine-tuning of layers with small $\Omega_i$ futile.

\begin{table}[!htbp]
\centering
\caption{Quantization results of ResNet20 on Cifar-10. We abbreviate quantization
bits used for weights as ``w-bits,'' activations as ``a-bits,''
testing accuracy as ``Acc,'' and compression ratio of weights/activations as ``W-Comp/A-Comp.''
Furthermore, we show results without using Hessian information (``Direct''), as well 
as other state-of-the-art methods~\cite{zhou2016dorefa,choi2018pact,Zhang_2018_ECCV}.
In particular, we compare with the recent DNAS approach of~\cite{wu2018mixed}.
Our method achieves similar testing performance with significantly higher compression (especially in activations).
 Here ``MP'' refers to mixed-precision quantization, where we report the lowest bits used for weights and activations.
Also note that \cite{zhou2016dorefa, choi2018pact, Zhang_2018_ECCV} use 8-bit for first and last layers.
The exact per-layer configuration for mixed-precision quantization of \OURS is presented in appendix.
}

\small
\setlength\tabcolsep{2.35pt}
\label{tab:resnet_compress_ratio}
\begin{tabular}{lccccccccccccc} \toprule
\ha Quantization & w-bits & a-bits & Acc    & W-Comp & A-Comp\\
    \midrule
\hc Baseline                           & 32     & 32 & 92.37      & 1.00$\times$    & 1.00$\times$  \\
\midrule
\ha Dorefa~\cite{zhou2016dorefa}       & 2      & 2  & 88.20      & 16.00$\times$  & 16.00$\times$\\
\ha Dorefa~\cite{zhou2016dorefa}       & 3      & 3  & 89.90      & 10.67$\times$  & 10.67$\times$\\
\ha PACT~\cite{choi2018pact}           & 2      & 2  & 89.70      & 16.00$\times$  & 16.00$\times$\\
\ha PACT~\cite{choi2018pact}           & 3      & 3  & 91.10      & 10.67$\times$  & 10.67$\times$\\

\ha LQ-Nets~\cite{Zhang_2018_ECCV}     & 2      & 2  & 90.20      & 16.00$\times$  & 16.00$\times$\\
\ha LQ-Nets~\cite{Zhang_2018_ECCV}     & 3      & 3  & 91.60      & 10.67$\times$  & 10.67$\times$\\
\ha LQ-Nets~\cite{Zhang_2018_ECCV}     & 2      & 32 & 91.80      & 16.00$\times$  & 1.00$\times$\\
\ha LQ-Nets~\cite{Zhang_2018_ECCV}     & 3      & 32 & 92.00      & 10.67$\times$  & 1.00$\times$\\

\ha DNAS~\cite{wu2018mixed}      & 1\MP      & 32  & {92.00}    & 16.60$\times$  & 1.00$\times$\\
\ha DNAS~\cite{wu2018mixed}      & 1\MP      & 32  & \bf{92.72} & 11.60$\times$  & 1.00$\times$\\
\midrule
\ha Direct                         & 2\MP      & 4  & 90.34      & 16.00$\times$  & 8.00$\times$\\
\hc \OURS                          & 2\MP      & 4  & \bf{92.22} & 13.11$\times$  & 8.00$\times$\\
\bottomrule 
\end{tabular}
\end{table}

\section{Results}
\label{sec:results}
In this section, we first present our quantization results for ResNet20 on Cifar-10, and then we present our results for Inception-V3, ResNet50, and SqueezeNext quantization on ImageNet.
See appendix for details regarding the training procedure and hyper-parameters used.

\paragraph{Cifar-10}
After computing the eigenvalues of block Hessian (shown in~\fref{fig:resnet_inception_eigs_surface}), we compute  the weighted sensitivity metric of~\eref{eqn:define_S}, along with $\Omega_i$ based on~\eref{eqn:define_O}.
We then perform the quantization based on \OURS algorithm.
Results are shown in~\tref{tab:resnet_compress_ratio}.

For comparison, we test the quantization performance without using the Hessian information, which we refer to as ``Direct'' method,
as well as other methods in the literature including Dorefa~\cite{zhou2016dorefa}, PACT~\cite{choi2018pact}, LQ-Net~\cite{Zhang_2018_ECCV}, and DNAS~\cite{wu2018mixed}, as shown in~\tref{tab:resnet_compress_ratio}.

For methods that use Mixed-Precision (MP), we report the lowest bits used for weights (``w-bits''), and activations (``a-bits'').

The Direct method achieves good compression, but it results in $2.03\%$ accuracy drop, as shown in~\tref{tab:resnet_compress_ratio}.%
% \footnote{Here, we are reporting the best numbers we achieved for
% this method after hyper-parameter tuning.}
Furthermore, comparison with other state-of-the-art  shows a similar trend.
There have been several methods proposed in the literature to address this reduction, with
the latest method introduced in~\cite{Zhang_2018_ECCV}, where a learnable
quantization method is used. As one can see, LQ-Nets results in $0.77\%$
accuracy degradation with $10.67\times$ compression ratio, whereas \OURS has only
$0.15\%$ accuracy drop with $13.11\times$ compression. Moreover, \OURS achieves
similar accuracy as compared to DNAS~\cite{wu2018mixed} but with $8\times$ higher compression ratio for activations.

\paragraph{ImageNet}
Here, we test the \OURS method for quantizing Inception-V3~\cite{szegedy2016rethinking} on ImageNet. 
Inception-V3 is appealing for efficient hardware implementation, as it does not 
use any residual connections. Such non-linear structures create dependencies that may
be very difficult to optimize for fast inference~\cite{yang2019synetgy}.
As before, we first compute the block Hessian eigenvalues, which are reported in~\fref{fig:resnet_inception_eigs_surface},
and then compute the corresponding weighted sensitivity metric. We also plot the 1D loss landscape of all Inception-V3 blocks in~\fref{fig:inception_1d_loss_appendix}.

We report the quantization results in~\tref{tab:inception_compress_ratio}, where
as before we compare with a direct quantization, as well as recently proposed ``Integer-Only''~\cite{jacob2018quantization}, and RVQuant methods~\cite{park2018value}.
Direct quantization of Inception-V3 (\textit{i.e.}, without use of second-order information),
results in $7.69\%$ accuracy degradation. 
Using the approach proposed in~\cite{jacob2018quantization} results in more than $2\%$ accuracy drop,
even though it uses higher bit precision.
However, \OURS results in a generalization gap of $2\%$ with a compression ratio
of $12.04\times$, both of which are better than previous work~\cite{jacob2018quantization,park2018value}.%
\footnote{We should emphasize here that the work of~\cite{jacob2018quantization} uses integer arithmetic,
and it is not completely fair to compare their results with ours.}

We also compare with Deep Compression~\cite{han2015deep} and the AutoML based method of HAQ, which has been recently introduced~\cite{wang2018haq}.
We compare our \OURS results with their
ResNet50 quantization, as shown in~\tref{tab:resnet50_compress_ratio}. \OURS achieves higher top-1 accuracy of 75.48\% with a model size of 7.96MB, whereas the AutoML based HAQ method has a top-1 of 75.30\% even with 16\% larger model size of 9.22MB.

Furthermore, we apply \OURS to quantize SqueezeNext~\cite{gholami2018squeezenext} on ImageNet. We choose
the wider SqueezeNext model which has a baseline accuracy of $69.38\%$ with 2.5 million parameters (10.1MB in single precision). We are able to quantize this model to uniform 8-bit precision, with just 0.04\% top-1 accuracy drop. Direct quantization of SqueezeNext (\textit{i.e.}, without use of second-order information), results in $3.98\%$ accuracy degradation. HAWQ results in an unprecedented 1MB model size, with only 1.36\% top-1 accuracy drop. The significance of this result is that it allows
deployment of the whole model on-chip or on hardwares with very limited memory and power constraints.

\begin{table}[!htbp]
\caption{
Quantization results of Inception-V3 on ImageNet. We abbreviate quantization
bits used for weights as ``w-bits,'' activations as ``a-bits,''
top-1 testing accuracy as ``Top-1,'' and weight compression ratio as``W-Comp.''
Furthermore, we compare \OURS with direct quantization method without using Hessian (``Direct'')
and Integer-Only~\cite{jacob2018quantization}. Here ``MP'' refers to mixed-precision quantization.
We report the exact per-layer configuration for mixed-precision quantization in appendix.
Compared to~\cite{jacob2018quantization,park2018value}, we achieve higher compression ratio with higher
testing accuracy.
}
\small
\setlength\tabcolsep{2.35pt}
\label{tab:inception_compress_ratio}
\centering
\begin{tabular}{lcccccccccccccc} \toprule

Method                                          & w-bits    & a-bits  & Top-1        &  W-Comp          & Size(MB)  
\\ % & A-Comp \\
\midrule
\hc  Baseline                                   & 32   & 32 & 77.45       & 1.00$\times$    &  91.2        \\ % & 1.00$\times$\\
\midrule
\ha  Integer-Only~\cite{jacob2018quantization}  & 8    & 8  & 75.40       & 4.00$\times$    &  22.8        \\ % & 4.00$\times$       \\
\ha  Integer-Only~\cite{jacob2018quantization}  & 7    & 7  & 75.00       & 4.57$\times$    &  20.0       \\ % & 4.57$\times$       \\
\ha  RVQuant~\cite{park2018value}  & 3\MP    & 3\MP  & 74.14       & 10.67$\times$    &  8.55        \\ % & 4.00$\times$       \\
\midrule
\ha  Direct                                 & 2\MP    & 4\MP  & 69.76       & 15.88$\times$   &  5.74     
\\ % & 8.00$\times$       \\ 
\hc \OURS                                   & 2\MP    & 4\MP  & \bf{75.52}  & 12.04$\times$   &  \bf{7.57}        \\ % & 8.00$\times$       \\
     \bottomrule 
\end{tabular}
\end{table}

\begin{table}[!htbp]
\caption{
Quantization results of ResNet50 on ImageNet.
We show results of state-of-the-art methods~\cite{zhou2016dorefa, choi2018pact, Zhang_2018_ECCV, han2015deep}.
In  particular,  we  also  compare  with  the  recent AutoML approach of~\cite{wang2018haq}.
Compared to~\cite{wang2018haq}, we achieve higher compression ratio with higher
testing accuracy. Also note that \cite{zhou2016dorefa, choi2018pact, Zhang_2018_ECCV} use 8-bit for first and last layers.
}
\small
\setlength\tabcolsep{2.35pt}
\label{tab:resnet50_compress_ratio}
\centering
\begin{tabular}{lcccccccccccccc} \toprule
Method                                          & w-bits    & a-bits  & Top-1        &  W-Comp          & Size(MB)  
\\ % & A-Comp \\
\midrule
\hc  Baseline                                   & 32   & 32 & 77.39       & 1.00$\times$    &  97.8        \\ % & 1.00$\times$\\
\midrule
\ha  Dorefa~\cite{zhou2016dorefa}  & 2    & 2  & 67.10       & 16.00$\times$    &  6.11        \\ % & 4.00$\times$       \\
\ha  Dorefa~\cite{zhou2016dorefa}  & 3    & 3  & 69.90       & 10.67$\times$    &  9.17       \\ % & 4.57$\times$       \\
\ha  PACT~\cite{choi2018pact}  & 2    & 2  & 72.20       & 16.00$\times$    &  6.11        \\ % & 4.00$\times$       \\
\ha  PACT~\cite{choi2018pact}  & 3    & 3  & 75.30       & 10.67$\times$    &  9.17       \\ % & 4.57$\times$       \\
\ha  LQ-Nets~\cite{Zhang_2018_ECCV}     & 3      & 3  & 74.20      & 10.67$\times$  & 9.17 \\ % & 10.67$\times$\\
\ha Deep Comp.~\cite{han2015deep}               & 3    & MP  & {75.10}  & 10.41$\times$  &  9.36        \\ % & 8.00$\times$   
\ha HAQ~\cite{wang2018haq}                 & MP   & MP  & {75.30}  & 10.57$\times$  &  9.22         \\ %& 8.00$\times$       \\
\midrule
\hc \OURS                                   & 2\MP    & 4\MP  & \bf{75.48}  & 12.28$\times$   &  \bf{7.96}        \\ % & 8.00$\times$       \\
% \midrule
     \bottomrule 
\end{tabular}
\end{table}

\begin{table}[!htbp]
\caption{
Quantization results of SqueezeNext on ImageNet.
We show a case where \OURS is used to achieved uniform quantization to 8 bits for both weights
and activations, with an accuracy similar to ResNet18.
We also show a case with mixed precision, where we compress SqueezeNext to a model with just 1MB size with only 1.36\% accuracy degradataion. Furthermore, we compare \OURS with direct quantization method without using Hessian (``Direct'').
}
\small
\setlength\tabcolsep{2.35pt}
\label{tab:squeezenext_compress_ratio}
\centering
\begin{tabular}{lcccccccccccccc} \toprule
Method              & w-bits    & a-bits  & Top-1        &  W-Comp          & Size(MB)  \\ % & A-Comp \\
\midrule
\hc  Baseline       & 32    & 32    & 69.38        & 1.00$\times$          &  10.1        \\ % & 1.00$\times$\\
\ha ResNet18~\cite{paszke2017automatic}      & 32    & 32    & 69.76       & 1.00$\times$     & 44.7\\                               
\midrule
\ha \OURS          & 8     & 8     & \bf{69.34}   & 4.00$\times$          &  2.53        \\ % & 8.00$\times$
\ha Direct      & 3\MP     & 8     & \bf{65.39}   & 9.04$\times$          &  1.12        \\
\hc \OURS       & 3\MP     & 8     &   \bf{68.02}           & 9.25$\times$          &  \bf{1.09}        \\
% \midrule
     \bottomrule 
\end{tabular}
\end{table}

% -----------------------------------------------------------------------
\begin{figure}[t]
\centering
\includegraphics[width=.49\textwidth]{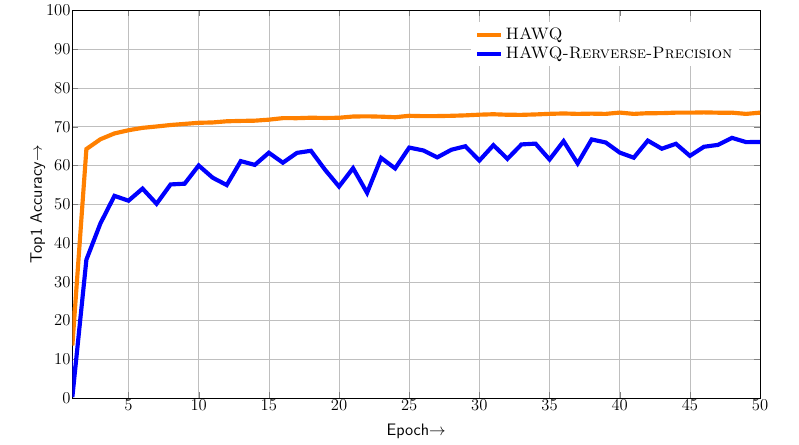}
\caption{
  Accuracy recovery from Hessian aware mixed precision quantization versus \OURS-Reverse-Precision quantization. Here, we show top-1 accuracy of quantized Inception-V3 on ImageNet. \OURS-Reverse-Precision achieves 66.72\% (compression-ratio 7.2) top-1 accuracy, while our \OURS method achieves 74.36\% (compression-ratio 12.0) top-1 accuracy (7.64\% better) with a higher convergence speed (30 epochs v.s. 50 epochs of \OURS-Reverse-Precision). 
  }
  \label{fig:mixedprecision}
\end{figure}
% -----------------------------------------------------------------------

% -----------------------------------------------------------------------
\begin{figure}[t]
\centering
\includegraphics[width=.49\textwidth]{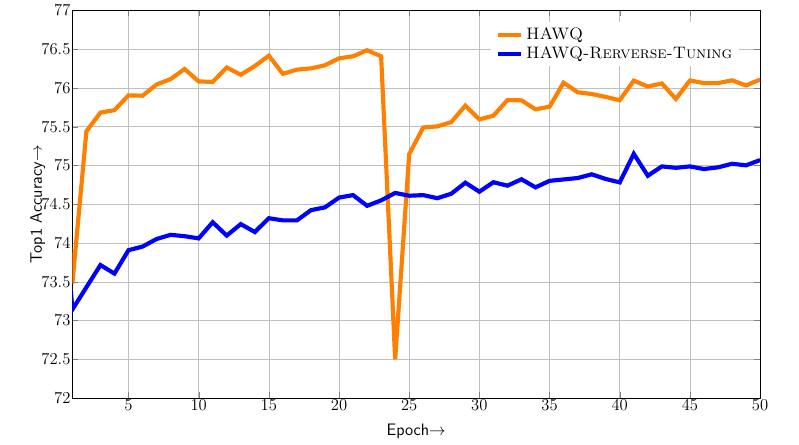}
\caption{
  Effectiveness of Hessian aware block-wise quantization. Here, \OURS shows the quantization process based on the descending order of $\Omega_i$ for Inception-V3 with Hessian aware quantization order. \OURS-Reverse-Tuning shows the quantization process of Inception-V3 with a reverse order. Note that \OURS finishes the fine-tuning of this block in just 25 epochs and switches to fine-tuning another block,
  whereas  \OURS-Reverse-Tuning takes 50 epochs for this block, before converging to sub-optimal top-1.
  }
  \label{fig:speedcomparison}
\end{figure}
% -----------------------------------------------------------------------

% --------------------------------------------

\section{Ablation Study}
\label{sec:ablation}

Here we discuss the ablation study for the \OURS.
The \OURS method has two main 
steps:
(i) relative precision order for different blocks using second-order information, and
(ii) relative order for fine-tuning these blocks.
Below we discuss the ablation study for each step separately.

\subsection{Hessian AWare Mixed Precision Quantization}
\label{subsec:ablation_mp}
We first discuss the ablation study for step (i), where the quantization precision is chosen
based on~\eref{eqn:define_S}. As discussed above, blocks with higher values of $S_i$ are 
assigned higher quantization precision, and vice versa for layers with relatively lower values of $S_i$.
For the ablation study
we reverse this order and avoid performing the block-wise fine-tuning of step (ii) so we can isolate
step (i). Instead of the fine-tuning phase, we re-train the whole network at once after the quantization is performed.
The results are shown in~\fref{fig:mixedprecision}, where we perform 50 epochs of fine-tuning using Inception-V3 on ImageNet. As one can see, \OURS results in significantly
better accuracy (74.26\% as compared to 66.72\%) than the reverse method (labeled as ``\OURS-Reverse-Precision'').
This is despite the fact that the latter approach only has a compression ratio of $7.2\times$, whereas \OURS has a compression
ratio of $12.0\times$.

Another interesting observation is that the convergence speed
of the Hessian aware approach is significantly faster than the reverse method.
Here, \OURS converges in about 30 epochs, whereas the \OURS-Reverse-Precision case takes 50 epochs before converging to a sub-optimal value (\fref{fig:mixedprecision}).

% ---------------------------------------------------------------------------
\subsection{Block-Wise Fine-Tuning}
\label{subsec:ablation_order}
Here we perform the ablation study for the Hessian based fine-tuning part of \OURS.
The block-wise fine tuning is performed based 
on $\Omega_i$ (\eref{eqn:define_O}) of each block.
The blocks are fine-tuned based on the descending order of $\Omega_i$.
Similar to the above, we compare the quantization performance when a reverse ordering is used (\textit{i.e.}, we use the
ascending order of $\Omega_i$ and refer to this as ``\OURS-Reverse-Tuning'').

We test this ablation study using Inception-V3 on ImageNet, as shown in~\fref{fig:speedcomparison}.
% For the first phase, \OURS starts fine tuning the block with the highest $\Omega_i$, whereas ``\OURS-Reverse-Tuning''
% starts fine-tuning the block with lowest $\Omega_i$. 
As one can see, the fine-tuning for \OURS method quickly converges in just 25 epochs,
allowing it to switch to fine-tuning the next block. However,
``\OURS-Reverse-Tuning'' takes more than 50 epochs to converge for this block.

\section{Conclusions}
\label{sec:conclusions}
We have introduced \OURS, a new quantization method for neural network training.
Our method is based on exploiting second-order (Hessian) information to systematically select both quantization precision as well as the order for block-wise fine-tuning.
We performed an ablation study for both the relative quantization bit-order
for different blocks, as well as the fine-tuning order.
We showed that \OURS can achieve good testing performance with high compression-ratio, as compared to state-of-the-art.
In particular, we showed results for ResNet20 on Cifar-10, where we can achieve similar testing performance as~\cite{wu2018mixed}, but with $8\times$ higher compression ratio for activations. We also showed results for Inception-V3 on ImageNet, for which we showed ultra low precision quantization results with 2-bit for weights and 4-bit for activations, with only $1.93\%$ accuracy drop. 
For ResNet50 model, our approach results in higher accuracy of 75.48\% with smaller model size of 7.96MB, as compared to HAQ method with top-1 of 75.30\% and 9.22MB~\cite{wang2018haq}.
Furthermore, our method applied to SqueezeNext can result in an unprecedented 1MB model size with 68.02 top-1 accuracy on ImageNet.

\emph{Limitations and Future Work.}
We believe it is critical for every work to clearly state its limitations, especially in this area. 
An important limitation is that
computing the second-order information adds some computational overhead. 
However,
we only need to compute the top eigenvalue of the Hessian, which can be found
using the matrix-free method presented in~\aref{alg:power_iteration}.
(The total computational overhead is equivalent to about 20 gradient back-propogations to compute
top Hessian eigenvalue of each block).
Another limitation is that in this work we solely focused on image classification, but it
would be interesting to see how \OURS would perform for more complex tasks such as
segmentation, object detection, or natural language processing.
Furthermore, one has to consider that implementation of a NN with mixed-precision inference
for embedded processors is not as straightforward as the case with
uniform quantization precision. Practical solutions have been proposed in recent works \cite{sharma2018bit}.
Another limitation is that we can only determine the relative ordering for quantization precision, and not
the absolute value of the bits. However, the search space for this is significantly smaller than the original exponential complexity.
Finally, even though we showed benefits of \OURS as compared to DNAS~\cite{wu2018mixed} or HAQ~\cite{wang2018haq}, it may be possible to combine these methods for more efficient AutoML search. We leave this as part of future work.

\section*{Acknowledgments}
This work was supported by a gracious fund from Intel corporation,
Berkeley Deep Drive (BDD), and Berkeley AI Research (BAIR) sponsors. We would like
to thank the Intel VLAB team for providing us with access to their computing cluster.
We also gratefully acknowledge the support of NVIDIA Corporation for their donation of two Titan Xp GPU used for this research.
We would also 
like to acknowledge ARO, DARPA, NSF, and ONR for providing partial support of this work.

{\small
\bibliographystyle{ieee}
\bibliography{ref}
}
\clearpage
\onecolumn

\section{Appendix}
Here, we provide additional experimental results as well as quantization
details for the neural networks that we tested.

\begin{itemize}

    \item In~\S~\ref{sec:details} we discuss the fine-tuning details.
    \item In~\S~\ref{sec:extra_results} we present extra results
    for 3D plots for loss landscape of different blocks of ResNet20 and Inception-V3 as well
    as exemplary results showing distribution of $\Omega_i$ in~\eref{eqn:define_O}.

    \item In~\S~\ref{sec:mixed_precision_detail} we show the exact bit-precision used for different
    blocks of ResNet20 on Cifar-10 as well as Inception-V3 on ImageNet.

\end{itemize}

\subsection{Fine-tuning details}
\label{sec:details}
 The results were tested on two classification datasets of Cifar-10 and ImageNet:

\paragraph{Cifar-10} This is a classification dataset with 10 classes
consisting of 50,000 training images and 10,000 test images
 of size $3\times32\times32$.
We used pre-trained ResNet20 model and performed quantization on this
model in PyTorch framework.
We follow the same learning rate policy as the baseline (\emph{i.e.}, decaying learning rate from 
0.1 to 0.0001).

\paragraph{ImageNet} This is a classification problem with 1000 classes consisting of more than 1.2 million training images 
and 50,000 validation images of size $3\times224\times224$ on SqueezeNext and ResNet50 , and $3\times299\times299$ on Inception-V3.
(i) We used pre-trained Inception-V3 model and used a fixed learning rate of 0.0002 for fine-tuning of each block.
(ii) We used pre-trained ResNet50 model and used a fixed learning rate of 0.0001 for fine-tuning of each block.
(iii) We used pre-trained SqueezeNext and used a fixed learning rate of 0.0001 for fine-tuning of each block.
All experiments were performed on PyTorch framework. As for data augmentation, we used standard random crop, resizing and horizontal flip in all experiments.

\subsection{Extra results}
\label{sec:extra_results}

\begin{figure*}[!htbp]
\centering
\includegraphics[width=.99\textwidth]{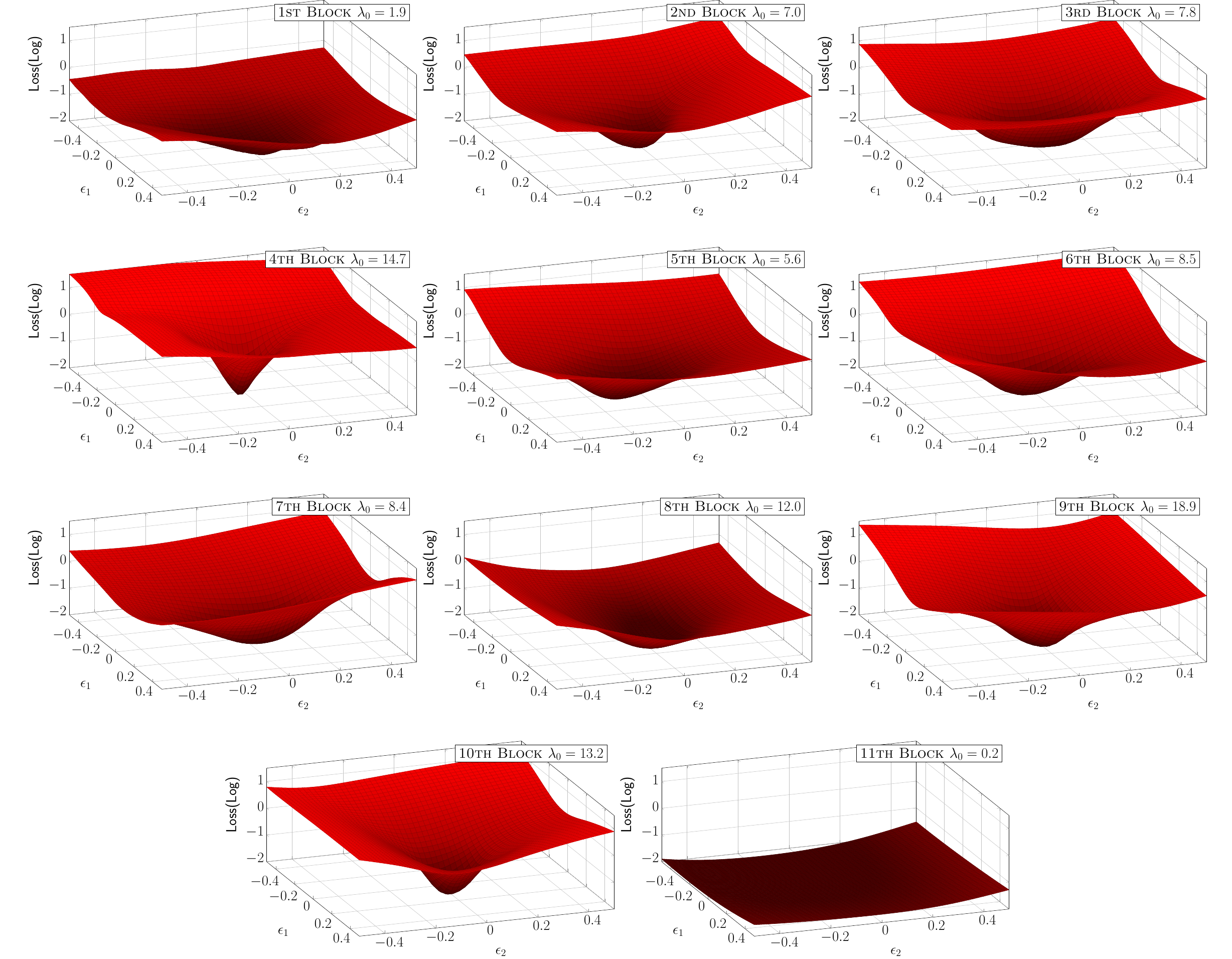}
\caption{
  3-D loss landscape of all blocks of ResNet20 on Cifar-10 along the first two dominant eigenvectors of the Hessian. Here $\epsilon_1$, $\epsilon_2$ are scalars that perturb the parameters of the corresponding block along the first and second dominant eigenvectors. 
  The corresponding eigenvalue distribution for different blocks is shown in~\fref{fig:resnet_inception_eigs_surface}.
  }
\label{fig:resnet20_surface_appendix}
\end{figure*}

\begin{figure*}[!htbp]
\centering
\includegraphics[width=.99\textwidth]{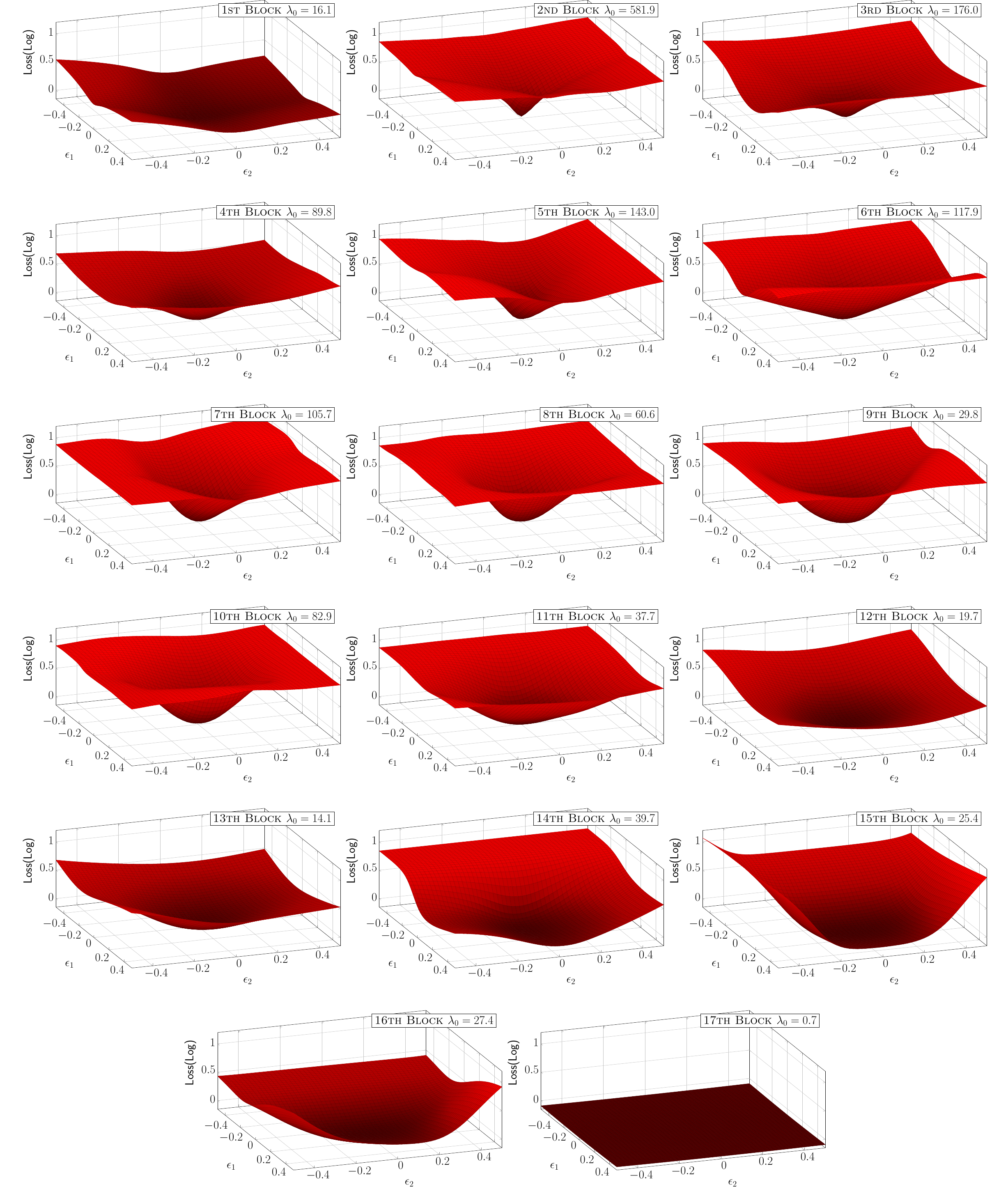}
\caption{
  3-D loss landscape of all blocks of InceptionV3 on ImageNet along the first two dominant eigenvectors of the Hessian. Here $\epsilon_1$, $\epsilon_2$ are scalars that perturb the parameters of the corresponding block along the first and second dominant eigenvectors. The corresponding eigenvalue distribution for different blocks is shown in~\fref{fig:resnet_inception_eigs_surface}.
  }
\label{fig:inception_surface_appendix}
\end{figure*}

In~\tref{tab:shift_table}, we show how $\Omega_i$ changes as a function of quantization precision. In~\fref{fig:resnet20_surface_appendix}, we plot the rest surface visualization of ResNet20 on Cifar-10. And in~\fref{fig:inception_surface_appendix}, we plot the rest surface visualization of Inception-V3 on ImageNet.

\begin{table}[!htbp]
\caption{Here we show how $\Omega_i$ changes as a function of 
target weight bit precision. Results are computed for ResNet20
on Cifar-10.}
% \small
% \setlength\tabcolsep{2.pt}
\label{tab:shift_table}
\centering
\begin{tabular}{lcccccccccccccc} \toprule
\diagbox[dir=SE]{Block}{Precision} & 8-bit & 6-bit & 4-bit & 3-bit & 2-bit\\
    \midrule
\ha  Block 3    & 0.03 & 0.52 & 9.25 & 41.9 & 191 \\
\ha  Block 5    & 0.05 & 0.81 & 14.0 & 65.1 & 309 \\
\ha  Block 8    & 0.29 & 4.83 & 84.8 & 392  & 2056 \\
     \bottomrule 
\end{tabular}
\end{table}

% ----------------------------------------------------------------

% ----------------------------------------------------------------
% ----------------------------------------------------------------

% ----------------------------------------------------------------

\subsection{Mixed-precision details}
\label{sec:mixed_precision_detail}
In this section, we give the details about how we separate blocks and details about weight/activation precision of each individual block. We show the exact bit-precision used for different
    blocks of ResNet20 (\tref{tab:resnet20_final_precision}) on Cifar-10 as well as Inception-V3 (\tref{tab:inception_final_precision}).

\begin{table}[!htbp]
\caption{Block seperation and final block precision of ResNet20 on Cifar-10. Here we abbreviate convolutional layer as ``Conv,'' fully connected layer as ``FC.''}
\small
\label{tab:resnet20_final_precision}
\centering
\begin{tabular}{lcccccccccccccc} \toprule
             &Block     &Layer(s)      &Layer Type    &Parameter Size  & Weight bit & Activation bit
             \\
    \midrule
             & Block 0  & Layer 0     & Conv          &      4.32e2     & 8          & 8 \\
    \midrule
\gc\gc             & Block 1  & Layer 1-2   & Conv          &      4.61e3     & 6           & 4  \\
    \midrule
             & Block 2  & Layer 3-4   & Conv          &      4.61e3     & 6           & 4  \\
    \midrule
\gc\gc             & Block 3  & Layer 5-6   & Conv          &      4.61e3     & 8           & 4  \\
             
    \midrule
             & Block 4  & Layer 7-8   & Conv          &      1.38e4     & 3           & 4   \\
             
    \midrule
\gc\gc             & Block 5  & Layer 9-10  & Conv          &      1.84e4     & 3          & 4   \\
             
    \midrule
             & Block 6  & Layer 11-12 & Conv          &      1.84e4     & 3          & 4   \\
             
    \midrule
\gc\gc             & Block 7  & Layer 13-14 & Conv          &      5.53e4     & 2         & 4   \\
             
    \midrule
             & Block 8  & Layer 15-16 & Conv          &      7.37e4     & 2          & 4   \\
             
    \midrule
\gc\gc             & Block 9  & Layer 17-18 & Conv          &      7.37e4     & 2         & 4   \\
             
    \midrule
             & Block 10 & Layer 19    & FC            &      6.40e2     & 3        & 8 \\
     \bottomrule 
\end{tabular}
\end{table}

\begin{table}[!htbp]
\caption{Block seperation and final block precision of  Inception-V3 on ImageNet. Here we abbreviate convolutional layer as ``Conv,'' fully connected layer as ``FC.''}
\small
\label{tab:inception_final_precision}
\centering
\begin{tabular}{lcccccccccccccc} \toprule
             &Block     &Layer(s)      &Layer Type    &Parameter Size(M)    & Weight bit   & Activation bit
             \\
    \midrule
             & Block 0  & Layer 0   & Conv         &    8.64e-4          &     6              & 6 \\
    \midrule
\gc             & Block 1  & Layer 1   & Conv         &    9.22e-3          &     6              & 6 \\
    \midrule
             & Block 2  & Layer 2   & Conv         &    1.84e-2          &     4              & 6 \\
    \midrule
\gc             & Block 3  & Layer 3   & Conv         &    5.12e-3          &     4              & 6 \\
    \midrule
             & Block 4  & Layer 4   & Conv         &    0.14             &     4              & 6 \\
    \midrule
\gc             & Block 5  & Layer 5-11   & Conv      &    0.25             &     4              & 4 \\
    \midrule
             & Block 6  & Layer 12-18  & Conv      &    0.28             &     4              & 4 \\
    \midrule
\gc             & Block 7  & Layer 19-25  & Conv      &    0.28             &     4              & 4 \\
    \midrule
             & Block 8  & Layer 26-29  & Conv      &    1.15             &     2              & 4 \\
    \midrule
\gc             & Block 9  & Layer 30-39  & Conv      &    1.29             &     4              & 4 \\
    \midrule
             & Block 10 & Layer 40-49  & Conv      &    1.69             &     4              & 4 \\
    \midrule
\gc             & Block 11 & Layer 50-59  & Conv      &    1.69             &     4              & 4 \\
    \midrule
             & Block 12 & Layer 60-69  & Conv      &    2.14             &     4              & 4 \\
    \midrule
\gc             & Block 13 & Layer 70-75  & Conv      &    1.70             &     2              & 4 \\
    \midrule
             & Block 14 & Layer 76-84  & Conv      &    5.04             &     2              & 4 \\
    \midrule
\gc             & Block 15 & Layer 85-93  & Conv      &    6.07             &     2              & 4 \\
    \midrule
             & Block 16 & Layer 94  & FC           &    2.05             &     2              & 4 \\
     \bottomrule

\end{tabular}
\end{table}

\end{document}